\documentclass[conference]{IEEEtran}
\IEEEoverridecommandlockouts
\usepackage{cite}
\usepackage{amsmath,amssymb,amsfonts}
\usepackage{algorithmic}
\usepackage[pdftex]{graphicx}
\usepackage{textcomp}
\usepackage{xcolor}  
\usepackage{booktabs}
\usepackage{makecell}
\usepackage{here}
\usepackage{url} 
\usepackage{multicol}
\usepackage{adjustbox} 
\usepackage{multirow}

\usepackage[linesnumbered,ruled,vlined]{algorithm2e}

\makeatletter
\newcommand{\newlineauthors}{%
  \end{@IEEEauthorhalign}\hfill\mbox{}\par
  \mbox{}\hfill\begin{@IEEEauthorhalign}
}
\makeatother

\def\BibTeX{{\rm B\kern-.05em{\sc i\kern-.025em b}\kern-.08em
    T\kern-.1667em\lower.7ex\hbox{E}\kern-.125emX}}

\usepackage{tikz} 
\usetikzlibrary{shapes.geometric, arrows, positioning, calc} 
\tikzstyle{data} = [
    rectangle, rounded corners,
    minimum width=2cm, minimum height=1cm,
    text centered, align=center,
    draw=black, fill=gray!20
]

\tikzstyle{signal_encoder} = [
    rectangle,
    minimum width=1.6cm, minimum height=1cm,
    text centered, align=center,
    draw=black, fill=blue!15
]
\tikzstyle{text_encoder} = [
    rectangle,
    minimum width=2.cm, minimum height=1cm,
    text centered, align=center,
    draw=black, fill=blue!15
]

\tikzstyle{attention} = [
    rectangle,
    minimum width=1.5cm, minimum height=1cm,
    text centered, align=center,
    draw=black, fill=green!20
]

\tikzstyle{process} = [
    rectangle,
    minimum width=2cm, minimum height=1cm,
    text centered, align=center,
    draw=black, fill=orange!20
]

\tikzstyle{arrow} = [
    thick, ->, >=stealth, draw=black
]

\begin{document}

\title{Retrieving Time-Series Differences Using Natural Language Queries}

\author{\IEEEauthorblockN{Kota Dohi, Tomoya Nishida, Harsh Purohit, Takashi Endo, Yohei Kawaguchi}
\IEEEauthorblockA{\textit{Research and Development Group, Hitachi, Ltd.}}
}

\maketitle

\begin{abstract}
Effectively searching time-series data is essential for system analysis; however, traditional methods often require domain expertise to define search criteria. Recent advancements have enabled natural language-based search, but these methods struggle to handle differences between time-series data. To address this limitation, we propose a natural language query-based approach for retrieving pairs of time-series data based on differences specified in the query. Specifically, we define six key characteristics of differences, construct a corresponding dataset, and develop a contrastive learning-based model to align differences between time-series data with query texts. Experimental results demonstrate that our model achieves an overall mAP score of 0.994 in retrieving time-series pairs.
\end{abstract}


\section{Introduction}
\label{sec:intro}
The state of any system can be represented as time-series data consisting of one or multiple channels. However, for system analysis, the most critical data often pertains to specific channels over limited time periods. Therefore, techniques for efficiently searching necessary data are essential~\cite{keogh2003need, kleist2015timeseries}.

Traditional time-series data search required expertise in defining search criteria and designing similarity measures~\cite{Christos1994, Keogh2001Dimensionality, nakamura2013shape}. This reliance on specialized knowledge made these methods inaccessible to non-experts, limiting their usability in broader contexts.
Recent studies have addressed this issue by using natural language queries for searching time-series data~\cite{Imani2019, ito2024clasp}. These methods allow users to retrieve time-series data by intuitively specifying language queries that describe the shape or trend of the data, such as ``The signal has an upward trend.'' However, a major limitation is their inability to search based on differences between multiple time-series, which are often essential in real-world applications.
For instance, in industrial applications, identifying deviations from normal operating conditions often requires comparing current sensor data to historical data. 
Consequently, existing language query methods are inadequate in practical applications.

To address this limitation, this study proposes a natural language query-based approach for retrieving pairs of time-series data with query-specified differences.
First, we define six key characteristics of differences between reference data and target data: upward trend, downward trend, spike, dropout, noise, and baseline. Based on these characteristics, we prepare pairs of time-series data and corresponding query texts, where each pair represents one of the defined characteristics.
Second, we develop a contrastive learning method to align the characteristics of differences between the reference and target data with query texts.
Finally, we evaluate the trained model on test data and examine its ability to retrieve pairs of reference and target data based on differences specified in natural language queries.

\section{Relation to prior work}
The task of retrieving specific time-series data from large datasets has been extensively studied in the field of data mining~\cite{Christos1994, Keogh2001Dimensionality, nakamura2013shape}. These studies typically used time-series data as the query and searched for time-series data with high similarity scores. However, defining an appropriate similarity metric requires careful consideration of the domain and use case, which often demands specialized knowledge. This reliance on expertise poses challenges for non-experts and limits the accessibility of these methods~\cite{Imani2019}.
To tackle this limitation, natural language-based search methods have been explored for time-series retrieval~\cite{Imani2019, ito2024clasp}. These methods allow users, including non-experts, to intuitively define search criteria by describing trends or patterns in human-interpretable terms. However, they cannot handle differences between multiple time-series. To address this, this study proposes a method for retrieving pairs of time-series data based on differences specified in the language query.



\section{Problem statement}
The reference data \( x_{\text{ref}} \) is time-series data that serves as the basis for comparison, while the target data \( x_{\text{tgt}} \) is the data being compared. 
The natural language query \( q \) describes the differences between the corresponding pairs of \( x_{\text{ref}} \) and \( x_{\text{tgt}} \).

Given the query \(q \), the task is to retrieve a set of corresponding pairs of \((x_{\text{ref}}, x_{\text{tgt}})\) from a set of \( N \) pairs. Formally, the objective is to find \( D_q \subseteq D \), where:
\[
D = \{ (x_{\text{ref}}^i, x_{\text{tgt}}^i) \mid i = 1, 2, \dots, N \},
\]
\[
D_q = \{ (x_{\text{ref}}^i, x_{\text{tgt}}^i) \in D \mid \text{matches descriptions in } q \}.
\]

To evaluate whether a pair \( (x_{\text{ref}}^i, x_{\text{tgt}}^i) \) matches the query \( q \), we compute a similarity score \( s_i \) based on the query and the pair's characteristics. The similarity score is defined as:
\[
s_i = f(x_{\text{ref}}^i, x_{\text{tgt}}^i, q),
\]
where \( f(\cdot) \) measures the extent to which the pair \( (x_{\text{ref}}^i, x_{\text{tgt}}^i) \) aligns with the descriptions provided in the query \( q \).


The pairs in \( D \) are ranked in descending order of their similarity scores \( s_i \). The retrieved set \( D_q \) consists of the top \( k \) pairs, formally defined as:
\[
D_q = \{ (x_{\text{ref}}^{i_j}, x_{\text{tgt}}^{i_j}) \mid j = 1, 2, \dots, k \},
\]
where \( i_1, i_2, \dots, i_k \) are the indices corresponding to the top \( k \) similarity scores \( s_{i_1}, s_{i_2}, \dots, s_{i_k} \).

\section{Preparation of Dataset}
\subsection{Key Characteristics of Differences between Time-Series}
\label{ssec:var_types}
There are numerous ways to describe the differences between two time-series. In this study, however, we focus on six key characteristics of differences: upward trend, downward trend, spike, dropout, noise, and baseline. These characteristics are selected based on their importance in capturing changes commonly observed in real-world time-series datasets.

\textbf{Upward / Downward trend:} 
Refers to the extent to which the data consistently increases or decreases over time. This represents the overall strength of the trend across all time points.

\textbf{Spike / Dropout:} Refers to the magnitude of short-term fluctuations. Variations in spikes or dropouts may reflect the severity of anomalies, sudden events, or transient disturbances within a system.

\textbf{Noise:} Refers to the level of random fluctuations in time-series data. Increased noise often indicates higher measurement uncertainty, sensor malfunctions, or unstable environmental conditions.

\textbf{Baseline:} Refers to the mean value of time-series data. Changes in baseline may result from sensor calibration errors, variations in operational conditions, or persistent deviations from expected values.

In this study, the reference data and target data pairs are assumed to differ in only one of these characteristics.




\subsection{Generation of Synthetic Time-Series Data Pairs}
\label{ssec:perturbation-func}
No publicly available dataset contains pairs of time-series data with differences in the characteristics defined in~\ref{ssec:var_types}. Therefore, we synthetically generate reference-target data pairs from real-world time-series data. 

We first normalize the data using min-max scaling to ensure that all values fall within the range \([0,1]\). Min-max scaling standardizes the data by uniformly adjusting its range, facilitating stable and efficient model training.

After normalization, we randomly select one of the six characteristics and apply the corresponding perturbation to the data.\footnote{If the upward trend is selected but the slope of the data is negative, the characteristic is switched to a downward trend. Similarly, if the downward trend is selected but the slope is positive, the characteristic is switched to an upward trend.}
The process of adding perturbation for each characteristic can be described as follows:

\textbf{Upward / Downward trend:} A linear trend is added to the data. 
Given a trend magnitude \( \alpha \) and original time-series data \(x_{\text{orig}}\), we define:
    \[
    x_{\text{ref/target}}(t) = 
    \begin{cases} 
        x_{\text{orig}}(t) + \alpha t, & \text{(Upward trend)} \\
        x_{\text{orig}}(t) - \alpha t. & \text{(Downward trend)}
    \end{cases}
    \]
    
To ensure that the perturbation does not result in a simple magnitude relationship between the reference and target data, we apply min-max scaling after adding the perturbation.

\textbf{Spike / Dropout:} A spike (sudden increase) or dropout (sudden decrease) is introduced at a randomly selected time \( t_s \). Given a magnitude \( \beta \), we define:
\[
x_{\text{ref/target}}(t) = 
\begin{cases} 
    x_{\text{orig}}(t) + \beta \delta(t - t_s), & \text{(Spike)} \\
    x_{\text{orig}}(t) - \beta \delta(t - t_s), & \text{(Dropout)}
\end{cases}
\]
where \( \delta(t - t_s) \) is the Dirac delta function at \( t_s \).

\textbf{Noise:} Random Gaussian noise with the noise level \( \gamma \) is added to the target data:
    \[
    x_{\text{ref/target}}(t) = x_{\text{orig}}(t) + \gamma v_{\text{noise}}(t),
    \]
    \[
    v_{\text{noise}}(t) \sim \mathcal{N}(0, 1).
    \]

\textbf{Baseline:} A constant baseline shift factor \( \theta \) is added to the entire time series:
    \[
    x_{\text{ref/target}}(t) = x_{\text{orig}}(t) + \theta.
    \]

\begin{table}[t]
\centering
\caption{Perturbation Parameters Used for Reference-Target Pair Generation}
\label{tab:variation_params}
\begin{tabular}{lcc}
    \toprule
    \textbf{Parameter name} & \textbf{\makecell{Smaller \\ perturbation level}} & \textbf{\makecell{Larger \\ perturbation level}} \\
    \midrule
    Trend magnitude \( \alpha \) & \( 0.0-0.5 \) & \( 0.5-1.0 \) \\
    Spike magnitude \( \beta \) & \( 0.0-0.1 \) & \( 0.1-0.5 \) \\
    Noise level \( \gamma \) & \( 0.0-0.05 \) & \( 0.05-0.1 \) \\
    Baseline shift \( \theta \) & \( 0.0-0.1 \) & \( 0.1-0.5 \) \\
    \bottomrule
\end{tabular}
\end{table}

To introduce various relative differences between the reference and target data, we prepare two levels of perturbation: a larger and a smaller perturbation level. The parameters (\( \alpha, \beta, \gamma, \theta \)) for these perturbations are randomly sampled within the ranges specified in Table~\ref{tab:variation_params}. These perturbations are applied to the same time-series data to generate two time-series, one of which is randomly assigned as the reference data and the other as the target data.

Depending on the characteristic of difference and whether the target data is subjected to a larger or smaller perturbation, there are 12 possible relationships between the reference and target data. Each reference-target pair is assigned a label $y_s$ from 1 to 12 to represent its relationship.


\begin{table}[]
\centering
\caption{Examples of queries and their associated reference-target data pairs. In the plots, the orange line represents the target data, and the blue line represents the reference data}
\renewcommand{\arraystretch}{1.5} 
\setlength{\tabcolsep}{6pt}       
\begin{tabular}{m{4cm} m{4cm}}
\hline
\begin{tabular}{r}\textbf{Query text}\end{tabular} & \textbf{Reference-target data pair} \\
\hline
The target data exhibits a stronger trend of increase than the reference data.&
\raisebox{-0.2\height}{\includegraphics[width=\linewidth]{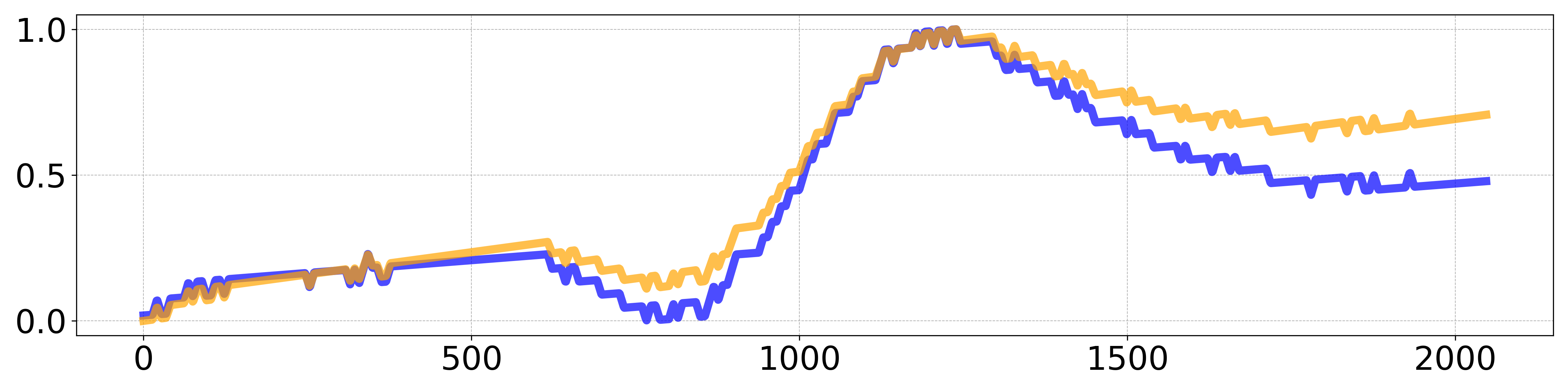}}
\\[2mm] \hline

The target dataset contains a significantly higher surge compared to the reference dataset.&
\raisebox{-0.2\height}{\includegraphics[width=\linewidth]{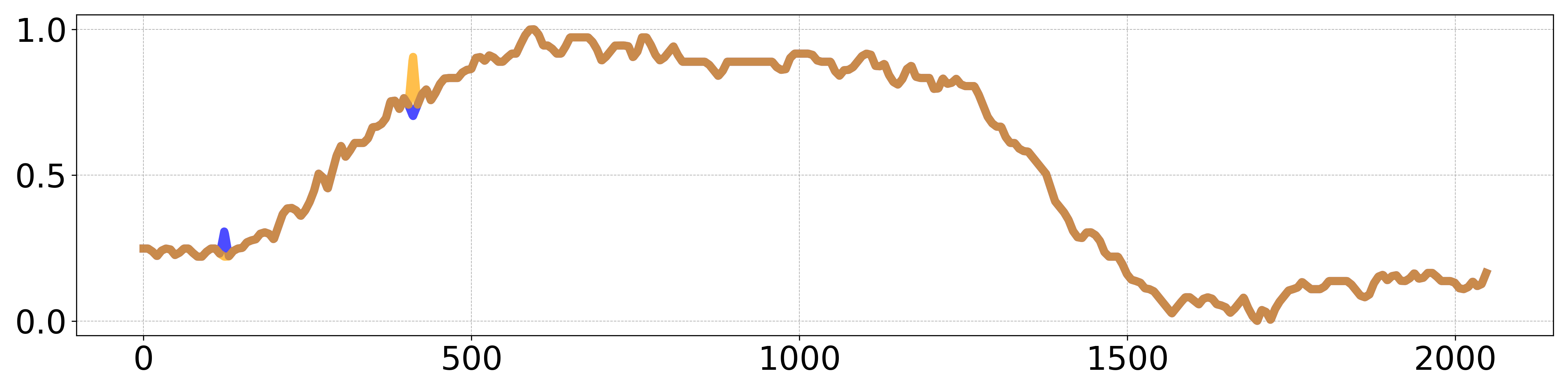}}
\\ \hline

The target dataset contains a greater degree of distortion relative to the reference dataset.&
\raisebox{-0.2\height}{\includegraphics[width=\linewidth]{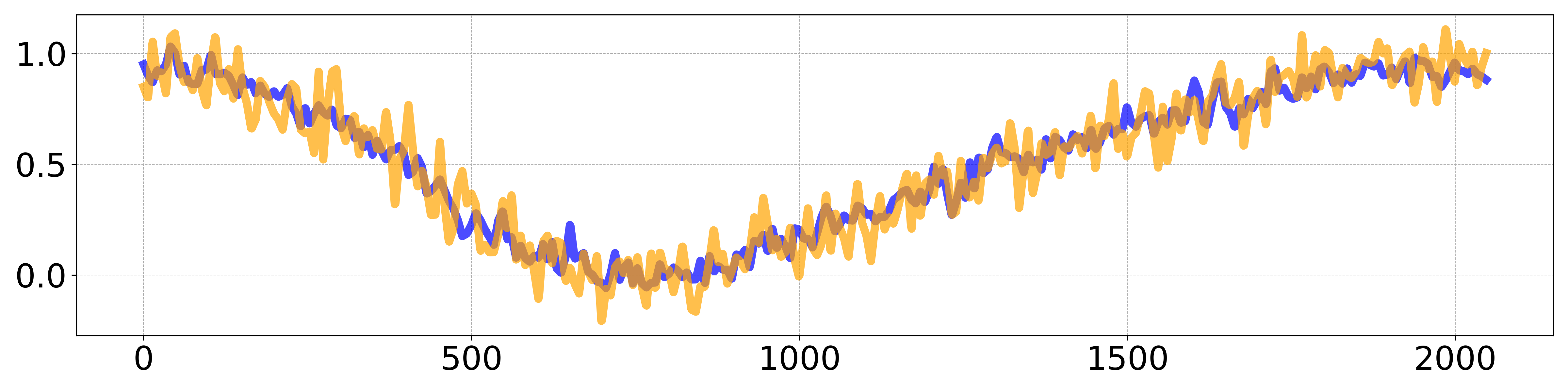}}
\\ \hline

The baseline present in the target dataset is noticeably higher than that of the reference dataset.& 
\raisebox{-0.2\height}{\includegraphics[width=\linewidth]{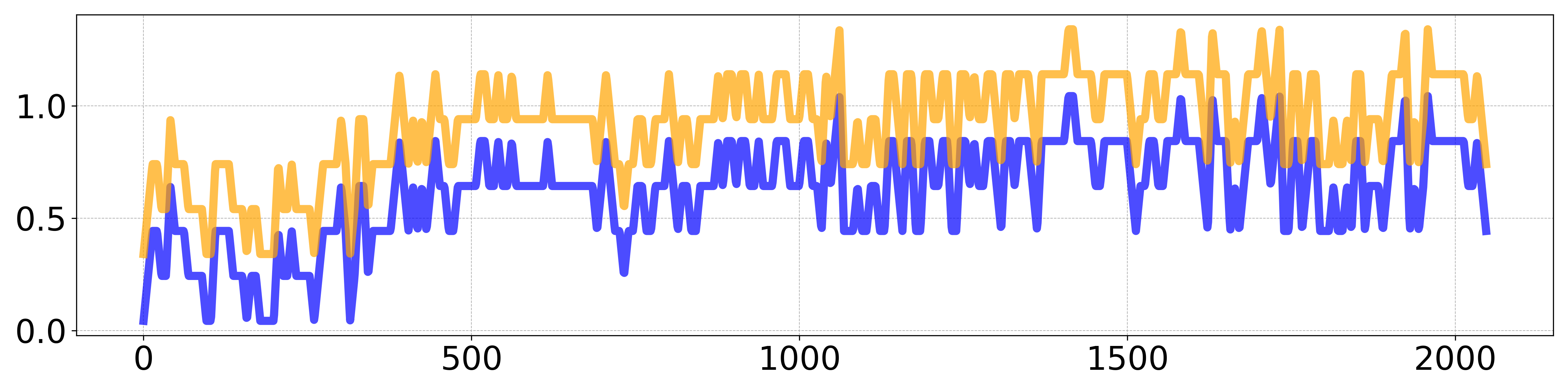}}
\\ \hline

\end{tabular}
\label{tab:spectrograms}
\end{table}


\subsection{Generation of Query Texts}
\label{ssec:gen_query}
We generate query texts for reference-target data pairs using the following template: 
\begin{quote}
    \textit{The target data has \{direction\} \{characteristic of difference\} than the reference data.}
\end{quote}
Here, \textit{\{direction\}} is either \textit{larger} or \textit{smaller}, indicating whether the target data is subjected to a larger or smaller perturbation. 
\textit{\{characteristic of difference\}} corresponds to one of the six predefined characteristics.

Once the template is instantiated with specific values, we rephrase the resulting texts using the Azure OpenAI API with the GPT-4o model~\cite{openai2023gpt4}. This process enhances the naturalness and diversity of the query texts while preserving their core semantic meaning.

Each query text is assigned a label $y_t$ according to the relationship between the reference and target data it describes, similar to the labeling method for reference-target data pairs. These labels are then used to generate the training dataset, consisting of reference-target data pairs and their corresponding query texts. Table \ref{tab:spectrograms} shows examples of query texts and their associated reference-target data pairs.


\begin{table*}[htbp]
    \centering
    \caption{\lowercase{m}AP scores for each relationship between the reference and target data. ``Total'' denotes the overall mAP score}
    \renewcommand{\arraystretch}{1.2} 
    \resizebox{1.0\linewidth}{!}{
    \begin{tabular}{>{\centering\arraybackslash}p{1.1cm}|>{\centering\arraybackslash}p{0.9cm}|>{\centering\arraybackslash}p{1.0cm}||>{\centering\arraybackslash}p{0.75cm}@{\hskip 3pt}>{\centering\arraybackslash}p{0.75cm}|>{\centering\arraybackslash}p{0.75cm}@{\hskip 3pt}>{\centering\arraybackslash}p{0.75cm}|>{\centering\arraybackslash}p{0.75cm}@{\hskip 3pt}>{\centering\arraybackslash}p{0.75cm}|>{\centering\arraybackslash}p{0.75cm}@{\hskip 3pt}>{\centering\arraybackslash}p{0.75cm}|>{\centering\arraybackslash}p{0.75cm}@{\hskip 3pt}>{\centering\arraybackslash}p{0.75cm}|>{\centering\arraybackslash}p{0.75cm}@{\hskip 3pt}>{\centering\arraybackslash}p{0.75cm}|>{\centering\arraybackslash}p{0.75cm}}
        \toprule
        \multicolumn{3}{c||}{\textbf{Method}} & 
        \multicolumn{12}{c|}{\textbf{Characteristic of difference}} & 
        \textbf{Total} \\
        \cmidrule{1-16}
        \multirow{2}{*}{\makecell[c]{\textbf{Signal}\\\textbf{encoder}}} & 
        \multirow{2}{*}{\makecell[c]{\textbf{Merge}\\\textbf{method}}} & 
        \multirow{2}{*}{\makecell[c]{\textbf{Cross}\\\textbf{attention}}} & 
        \multicolumn{2}{c|}{\textbf{Upward}} & 
        \multicolumn{2}{c|}{\textbf{Downward}} & 
        \multicolumn{2}{c|}{\textbf{Spike}} & 
        \multicolumn{2}{c|}{\textbf{Dropout}} & 
        \multicolumn{2}{c|}{\textbf{Noise}} & 
        \multicolumn{2}{c|}{\textbf{Baseline}} & \\
        \cmidrule{4-16}
        & & & 
        \textbf{larger} & \textbf{smaller}
        & \textbf{larger} & \textbf{smaller}
        & \textbf{larger} & \textbf{smaller}
        & \textbf{larger} & \textbf{smaller}
        & \textbf{larger} & \textbf{smaller}
        & \textbf{larger} & \textbf{smaller} & \textbf{} \\
        \midrule
        1D-CNN & concat & & 0.770 & 0.610 & 0.562 & 0.659 & 0.687 & 0.697 & 0.522 & 0.497 & \textbf{1.000} & 0.995 & \textbf{1.000} & 0.969 & 0.747 \\
        1D-CNN & diff & & 0.448 & 0.494 & 0.637 & 0.443 & 0.912 & 0.969 & 0.983 & 0.956 & \textbf{1.000} & \textbf{1.000} & \textbf{1.000} & 0.969 & 0.818 \\
        1D-CNN & concat & \checkmark & \textbf{1.000} & 0.991 & \textbf{1.000} & \textbf{1.000} & 0.613 & 0.491 & 0.471 & 0.516 & 0.987 & 0.956 & \textbf{1.000} & \textbf{1.000} & 0.835 \\
        1D-CNN & diff & \checkmark & 0.865 & 0.872 & 0.860 & 0.794 & 0.634 & 0.321 & 0.358 & 0.456 & 0.966 & 0.911 & \textbf{1.000} & \textbf{1.000} & 0.753 \\
        Informer & concat & & \textbf{1.000} & \textbf{1.000} & \textbf{1.000} & \textbf{1.000} & 0.815 & 0.768 & 0.847 & 0.745 & \textbf{1.000} & \textbf{1.000} & 0.993 & \textbf{1.000} & 0.931 \\
        Informer & diff & & \textbf{1.000} & \textbf{1.000} & \textbf{1.000} & \textbf{1.000} & \textbf{0.975} & \textbf{0.974} & \textbf{1.000} & \textbf{0.980} & \textbf{1.000} & \textbf{1.000} & \textbf{1.000} & \textbf{1.000} & \textbf{0.994} \\
        Informer & concat & \checkmark & \textbf{1.000} & \textbf{1.000} & \textbf{1.000} & \textbf{1.000} & 0.969 & 0.956 & \textbf{1.000} & 0.962 & \textbf{1.000} & \textbf{1.000} & 0.993 & \textbf{1.000} & 0.990 \\
        Informer & diff & \checkmark & \textbf{1.000} & 0.968 & \textbf{1.000} & \textbf{1.000} & 0.973 & 0.961 & 0.994 & 0.988 & \textbf{1.000} & \textbf{1.000} & \textbf{1.000} & \textbf{1.000} & 0.990 \\
        \bottomrule
    \end{tabular}
    }
    \label{tab:map_scores}
\end{table*}

\section{Model for aligning time-series data and query texts}
\label{sec:model-architecture}
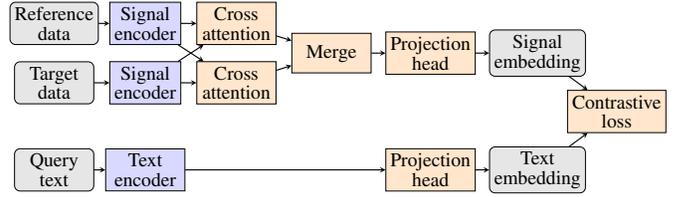
\begin{figure}[t]
\centering
\tikzset{font=\fontsize{14}{14}\selectfont} 
\resizebox{1.0\linewidth}{!}{
\begin{tikzpicture}[node distance=1.5cm and 1.5cm]
    \node (ref) [data] {Reference \\data};
    \node (tgt) [data, below of=ref] {Target \\data};
    \node (query) [data, below of=tgt, yshift=-0.7cm] {Query \\text};
    
    \node (encoder1) [signal_encoder, right of=ref, xshift=0.8cm] {Signal \\encoder};
    \node (encoder2) [signal_encoder, right of=tgt, xshift=0.8cm] {Signal \\encoder};
    \node (t5) [text_encoder, right of=query, xshift=0.8cm] {Text \\encoder};

    \node (crossattn1) [process, right of=encoder1, xshift=0.8cm] {Cross \\attention};
    \node (crossattn2) [process, right of=encoder2, xshift=0.8cm] {Cross \\attention};
    \node (diff) [process, xshift=2.4cm] at ($(crossattn1)!0.5!(crossattn2)$) {Merge};
    \node (proj_diff) [process, right of=diff, xshift=1cm] {Projection \\head};
    \node (z_diff) [data, right of=proj_diff, xshift=1.2cm]{Signal \\ embedding};
    \node (proj_text) [process] at (t5 -| proj_diff) {Projection \\head};

    \node (z_text) [data] at (t5 -| z_diff) {Text \\embedding};

    \node (contrast) [process, xshift=2cm] at ($(z_diff)!0.5!(z_text)$){Contrastive \\loss};

    \draw [arrow, thick] (ref) -- (encoder1);
    \draw [arrow, thick] (tgt) -- (encoder2);
    \draw [arrow, thick] (encoder1) -- (crossattn1);
    \draw [arrow, thick] (encoder1) -- (crossattn2);
    \draw [arrow, thick] (encoder2) -- (crossattn2);
    \draw [arrow, thick] (encoder2) -- (crossattn1);
    
    \draw [arrow, thick] (crossattn1) -- (diff);
    \draw [arrow, thick] (crossattn2) -- (diff);
    \draw [arrow, thick] (diff) -- (proj_diff);
    \draw [arrow, thick] (proj_diff) -- (z_diff);
    \draw [arrow, thick] (z_diff) -- (contrast);
    
    \draw [arrow, thick] (query) -- (t5);
    \draw [arrow, thick] (t5) -- (proj_text);
    \draw [arrow, thick] (proj_text) -- (z_text);
    \draw [arrow, thick] (z_text) -- (contrast);

\end{tikzpicture}
}
\caption{Overview of the model architecture for aligning time-series data with query texts.}
\label{fig:model_architecture}
\end{figure}

Fig.~\ref{fig:model_architecture} illustrates the overview of our model for aligning time-series data with query texts. The model encodes reference-target data pairs and query texts into a shared representation space and leverages contrastive learning for alignment. The framework comprises three key components: signal and text encoders, a method for merging signal embeddings, and supervised contrastive learning across time-series and text data.

\subsection{Encoding Pairs of Time-Series Data and the Query Texts}
First, the reference data \( x_{\text{ref}} \) and target data \( x_{\text{tgt}} \) are encoded into \(z_{\text{ref}}\) and \(z_{\text{tgt}}\) using a common signal encoder.  
Optionally, cross-attention can be applied between \( z_{\text{ref}} \) and \( z_{\text{tgt}} \) to effectively model the relative correlations between the reference and target data.  
The \(z_{\text{ref}}\) and \(z_{\text{tgt}}\) are then merged and passed into the projection head, which adjusts its dimensionality to align with that of the query texts.  
We prepare two methods for merging \(z_{\text{ref}}\) and \(z_{\text{tgt}}\): taking their difference (\textit{diff}) and concatenating them along the embedding dimension (\textit{concat}).

The query texts are also encoded using a text encoder, and the embedding is passed into the projection head.

\subsection{Supervised Contrastive Learning with Relationship Labels}
\label{ssec:contrastive-label}
Let us denote the signal and text embeddings obtained from a batch of $N$ samples by
\[
Z_{\text{signal}} 
= 
\bigl[z_{s1}, \,z_{s2}, \,\ldots, \,z_{sN}\bigr]
\quad 
\in \mathbb{R}^{N \times d},
\]
\[
Z_{\text{text}}
= 
\bigl[z_{t1}, \,z_{t2}, \,\ldots, \,z_{tN}\bigr]
\quad 
\in \mathbb{R}^{N \times d}.
\]
Here, $d$ denotes the embedding dimensionality. Each $z_{si}$ corresponds to the embedding of the $i$-th reference-target data pair, while each $z_{ti}$ corresponds to the embedding of the $i$-th query text.

Both the signal and text embeddings are first normalized using L2 normalization.
Then, we compute a logit matrix $M \in \mathbb{R}^{N \times N}$ whose entries measure the similarity between text and signal embeddings. Specifically,
\[
M_{ij} 
\;=\; 
(z_{ti} \,\cdot\, z_{sj}) / {\tau},
\]
where $\tau > 0$ is a temperature parameter that controls the scale of the logits, and $z_{ti} \cdot z_{sj}$ denotes the dot product of the $i$-th text embedding and the $j$-th signal embedding.

Based on the labels for each reference-target data pair $(y_{si})$ and each query text $(y_{ti})$, we construct a target matrix
$G \;\in\; \mathbb{R}^{N \times N}$
such that
\[
G_{i,j} 
\;=\; 
\begin{cases}
1 & \text{if } y_{ti} = y_{sj}, \\
0 & \text{otherwise}.
\end{cases}
\]
Viewed row-by-row, this matrix captures which reference-target data pairs match which query texts. 
We then normalize each row by dividing all elements in that row by their collective sum, which ensures that each row has values adding up to 1.
Hence, each row can be interpreted as a probability distribution over signal embeddings.

Let 
$\arg\max\bigl(G_{i,:}\bigr)$ denote the index of the largest entry in the $i$-th row of $G$. We define two cross-entropy losses. The first treats each row of $M$ as a predictive distribution over signal embeddings for a given text embedding:
\[
L_\text{texts} 
\;=\; 
\mathrm{CE}\Bigl(M, \;\arg\max(G)\Bigr),
\]
while the second treats each column (i.e., $M^\top$) as a predictive distribution over text embeddings for a given signal embedding:
\[
L_\text{signals}
\;=\;
\mathrm{CE}\Bigl(M^\top, \;\arg\max(G)\Bigr).
\]
Here, $\mathrm{CE}(\cdot,\cdot)$ denotes the standard cross-entropy loss function, computed along rows of the logit matrix (or its transpose).

Finally, we average these two losses to obtain the overall contrastive objective:
\[
L_\text{contrastive}
\;=\;
\frac{L_\text{texts} + L_\text{signals}}{2}.
\]

\section{Experiments}
\subsection{Dataset}
We used the TACO dataset~\cite{dohi2024domain} for generating pairs of reference data and target data. The TACO dataset consists of 1.2 million sensor time-series. First, we linearly interpolated each time-series so that all time-series have 2,048 points. Each time-series was then normalized using min-max scaling.  
Next, we extracted time-series one by one and
created pairs of reference data, target data, and their corresponding labels, as described in~\ref{ssec:perturbation-func}. In total, we generated 100,000 pairs for the training dataset, 2,000 pairs for the validation dataset, and 400 pairs for the test dataset.

The query texts were generated following the procedure described in~\ref{ssec:gen_query}. For each of the 12 relationships between reference and target data, we created 1,000 unique query texts. These query texts were then randomly divided into 900 for the training dataset and 100 for the test. The label was also assigned for each query text.
For each pair of reference-target data and its label, we randomly selected one of the query text with the same label and assigned it to the pair.

\subsection{Experimental conditions}
For the signal encoder, we employed 1D-CNN~\cite{lecun1998, wang2017} and Informer~\cite{zhou2021informer}. For the text encoder we used BART-Large-XSum~\cite{lewis2020bart, narayan2018xsum} \footnote{\url{https://huggingface.co/facebook/bart-large-xsum}}, T5-base~\cite{raffel2020exploring}\footnote{https://huggingface.co/google-t5/t5-base}, and RoBERTa-large~\cite{liu2019roberta}\footnote{https://huggingface.co/FacebookAI/roberta-large}. The cross attention module was implemented using a multi-head attention mechanism with eight attention heads. The projection head is a multi-layer perceptron consisting of two linear layers with a GELU activation, dropout, a residual connection, and layer normalization.
During training, we set the batch size to 512. The learning rates were set as follows: \( 1\times10^{-3} \) for the projection head, \( 1\times10^{-5} \) for the cross attention module and signal encoder, and \( 1\times10^{-4} \) for the text encoder.
We trained the model for 100 epochs and saved the version that achieved the minimum validation loss. The temperature parameter for contrastive learning was set to 1.0.

\subsection{Evaluation}
To evaluate the retrieval performance of the model we described in~\ref{sec:model-architecture}, we used the mAP score~\cite{everingham2010pascal}.
First, for each relationship between the reference and target data, we converted 100 test query texts into text embeddings. We also converted 400 test pairs of reference and target data into signal embeddings. Then, for each query's text embedding, we computed the cosine similarities with the signal embeddings. Using the obtained similarities and the labels assigned to each reference-target data pair and query text, we calculated the mAP score for each of the twelve relationships.
We also calculated the overall mAP score across all relationships.



\subsection{Results}
We first conducted experiments with eight different configurations of signal encoders, merge methods, and cross-attention. In these experiments, the Bart-Large-XSum was used as the text encoder. During training, the signal encoder, cross-attention modules, and projection heads were trained from scratch, and the text encoder's parameters were unfrozen for fine-tuning.
Table~\ref{tab:map_scores} presents the mAP scores for each type of relationships between reference and target data. 
The highest overall mAP score of 0.994 was achieved with the Informer, the \textit{diff} merge method, and without cross-attention. Other configurations using the Informer with cross-attention also achieved comparable performance. The Informer consistently outperformed the 1D-CNN across most configurations, which can be attributed to its stronger ability to model long-term dependencies in sequences. Unlike characteristics such as \textit{noise} or \textit{baseline}, differences in \textit{upward / downward trend} cannot be captured solely through local relationships. This limitation likely contributed to the lower performance of the standalone 1D-CNN. However, the application of cross-attention to the 1D-CNN significantly improved its performance on \textit{upward / downward trend}, highlighting the importance of mechanisms that account for global dependencies.
The \textit{diff} merge method excelled in retrieving \textit{spike} and \textit{dropout}, particularly when used without cross-attention. This may be attributed to the fact that computing the differences between the embeddings of the reference and target data effectively emphasizes the unique features of the differences. The cross-attention improved the overall performance for the \textit{concat} merge method. However, its impact on other configurations was less consistent.

\begin{table}[t]
\centering
\caption{Overall mAP scores with different text encoders, frozen or unfrozen text encoder parameters, and supervised or self-supervised contrastive learning}
\resizebox{1.0\linewidth}{!}{
\begin{tabular}{>{\raggedright\arraybackslash}p{2.3cm}||>{\centering\arraybackslash}p{1.8cm}>{\centering\arraybackslash}p{1.8cm}>{\centering\arraybackslash}p{1.3cm}>{\centering\arraybackslash}p{1.5cm}}
\toprule
\textbf{Text encoder} 
& \makecell[c]{\textbf{Frozen /}\\\textbf{self-supervised}}
& \makecell[c]{\textbf{Unfrozen /}\\\textbf{self-supervised}}
&\makecell[c]{\textbf{Frozen /}\\\textbf{supervised}} 
& \makecell[c]{\textbf{Unfrozen /}\\\textbf{ supervised}\\\textbf {(Ours)}} \\
\midrule
BART-Large-XSum  & 0.208 & 0.226 & 0.986& \textbf{0.994}\\
T5-base    & 0.193& 0.210& 0.990& \textbf{0.992}\\
RoBERTa-large     & 0.214 & 0.208& 0.989& \textbf{0.994}\\
\bottomrule
\end{tabular}
}
\label{tab:text_encoder_comparison}
\end{table}


We also conducted experiments with various text encoders to compare their performance. In addition, we explored configurations combining self-supervised contrastive learning with frozen text encoder parameters. This was motivated by our previous works on contrastive learning between time-series data and text~\cite{dohi2024domain, ito2024clasp}, where the text encoder parameters were frozen during training, and the self-supervised contrastive learning framework proposed in~\cite{radford2021learningtransferablevisualmodels} was employed.

\begin{figure}[t]
    \centering
    \includegraphics[width=\linewidth]{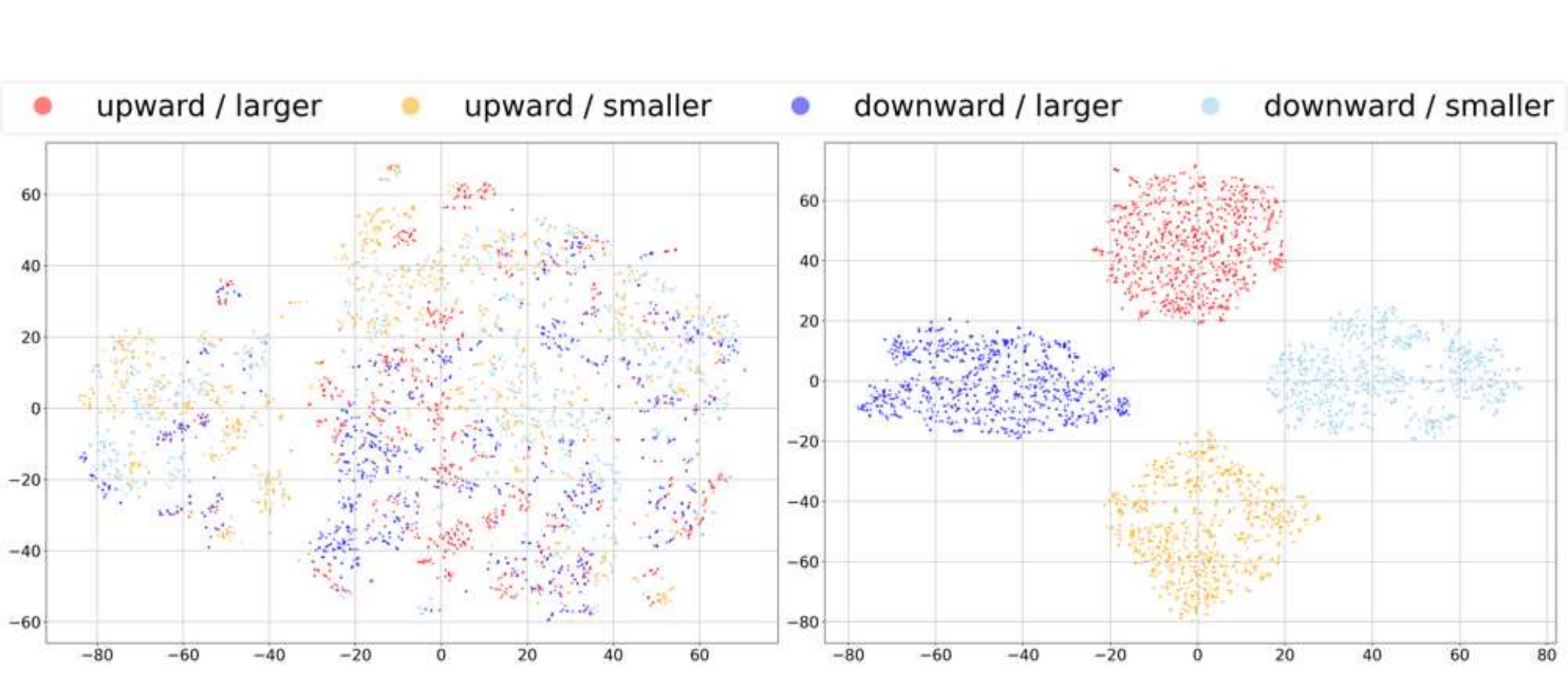}
    \caption{t-SNE plots of query text embeddings. \textbf{Left:} Embeddings generated by the pre-trained BART-Large-XSum model. \textbf{Right:} Embeddings generated by the fine-tuned BART-Large-XSum model.}
    \label{fig:text_emb_figs}
\end{figure}

Table~\ref{tab:text_encoder_comparison} summarizes the overall mAP score for each configuration. In these experiments, the signal encoder was set to Informer, the merge method was \textit{diff}, and cross-attention was not employed. The term ``supervised'' refers to the supervised contrastive learning framework described in~\ref{ssec:contrastive-label}, while ``self-supervised'' denotes self-supervised contrastive learning, which does not utilize labels. Across all text encoders, the ``unfrozen / supervised'' configuration consistently achieved the highest performance. Conversely, the performance of self-supervised configurations was generally lower.
Fig.~\ref{fig:text_emb_figs} presents t-SNE plots of 4,000 query text embeddings representing four different relationships. The left-hand side plot, which depicts text embeddings from the pre-trained BART-Large-XSum model, fails to distinguish embeddings corresponding to different relationships. In contrast, the right-hand side plot, generated using BART-Large-XSum fine-tuned with the ``unfrozen / supervised'' configuration, successfully separates embeddings for all relationships. These plots illustrate that pre-trained text encoders are not well-optimized for capturing nuanced differences in relationships, which led to the failure of self-supervised contrastive learning.

\section{Conclusion}
We proposed a natural language query-based method for retrieving pairs of time-series data with specified differences. We first created pairs of time-series and query texts based on six key characteristics of differences. We then developed contrastive learning methods for aligning differences in time-series data with query texts.  
Evaluation with test data showed overall mAP score of 0.994, confirming accurate retrieval across all characteristics of differences. 

\bibliographystyle{IEEEtran}
\bibliography{refs}
\end{document}